\documentclass{article}
\usepackage{PRIMEarxiv}
\usepackage[utf8]{inputenc}
\usepackage[T1]{fontenc}
\usepackage{hyperref}
\usepackage{url}
\usepackage{booktabs}
\usepackage{amsfonts}
\usepackage{nicefrac}
\usepackage{microtype}
\usepackage{lipsum}
\usepackage{fancyhdr}
\usepackage{graphicx}
\graphicspath{{media/}}
\usepackage{xcolor}
\usepackage{caption}

\usepackage{algorithm}
\usepackage{algpseudocode}
\usepackage{amsmath}
\usepackage{amsfonts}
\usepackage{amssymb}

\usepackage{amsmath}
\usepackage{graphicx}
\usepackage{subfig}

\title{Negotiated Representations to Prevent Forgetting in Machine Learning Applications 
\thanks{\textit{\underline{Citation}}: 
\textbf{GitHub Link https://github.com/nurikorhan/Negotiated-Representations-for-Continual-Learning}} 
}

\author{
  Nuri Korhan \\
  Istanbul Technical University  \\
  Maslak, Istanbul, Turkey \\
  \texttt{korhan@itu.edu.tr} \\
  \And
  Ceren Öner \\
   Istanbul Technical University \\
  Maslak, Istanbul, Turkey\\
  \texttt{csalkin@itu.edu.tr} \\
}

\begin{document}
\maketitle

\section{Abstract}
Catastrophic forgetting is a significant challenge in the field of machine learning, particularly in neural networks. When a neural network learns to perform well on a new task, it often forgets its previously acquired knowledge or experiences. This phenomenon occurs because the network adjusts its weights and connections to minimize the loss on the new task, which can inadvertently overwrite or disrupt the representations that were crucial for the previous tasks. As a result, the network's performance on earlier tasks deteriorates, limiting its ability to learn and adapt to a sequence of tasks.

In this paper, we propose a novel method for preventing catastrophic forgetting in machine learning applications, specifically focusing on neural networks. Our approach aims to preserve the network's knowledge across multiple tasks while still allowing it to learn new information effectively. We demonstrate the effectiveness of our method by conducting experiments on various benchmark datasets, including Split MNIST, Split CIFAR-10, Split Fashion MNIST, and Split CIFAR-100. These datasets are created by dividing the original datasets into separate, non-overlapping tasks, simulating a continual learning scenario where the model needs to learn multiple tasks sequentially without forgetting the previous ones.

Our proposed method tackles the catastrophic forgetting problem by incorporating negotiated representations into the learning process, which allows the model to maintain a balance between retaining past experiences and adapting to new tasks. By evaluating our method on these challenging datasets, we aim to showcase its potential for addressing catastrophic forgetting and improving the performance of neural networks in continual learning settings.
\section{Introduction}
The problem of catastrophic forgetting (the loss or disruption of previously learned information when new information is learned) in neural networks is commonly faced in deep learning tasks due to selected optimizers that are relied on Gradient Descent algorithm. 
\cite{rebuffi2017icarl,hou2019learning,kang2022class}. Decreasing the number of epochs and exploring rehearsal mechanisms (the retraining of some of the previously learned information as the new information is added) is critical for extraction of a potential solution.This point comes from the idea of the “stability / plasticity dilemma” \cite{grossberg1987competitive,carpenter1988art}, and emphasis on the representations developed by neural networks should be plastic enough to change to adapt to changing environments and learn new things, but stable enough so that important information is preserved over time. The dilemma is that while both are desirable properties, the requirements of stability and plasticity are in conflict. Stability depends on preserving the structure of representations, plasticity depends on altering it. An appropriate balance is difficult to achieve.

While stability / plasticity issues are very general, the term “catastrophic forgetting”
has tended to be associated with a specific class of networks, namely static networks
employing supervised learning. This broad class includes probably the majority of
commonly used and applied networks (such as the very influential back-propagation
family and Hopfield nets). Other types of network – for example
dynamic / “constructive” networks and unsupervised networks – are not necessarily
prone to catastrophic forgetting as it is typically described in this context. Dynamic
networks are those that use learning mechanisms where the number of units and
connections in the network grows or shrinks in response to the requirements of
learning \cite{hertz1991introduction} – new units can
be added to encode newly learned information without disrupting existing units. In
unsupervised networks there are no “correct” / target outputs to be forgotten. More
general stability / plasticity issues certainly arise, but may be more amenable to
solution in the unsupervised framework – \cite{gillies1991stability} provides valuable information of these issues and explores possible methods.

Continual Learning focuses on the problem of learning from
an infinite piece of data, with the goal of incrementally spreading extracted outcomes and using this outcomes for further learning phases \cite{chen2018continual}. The critical point in learning process is that a small portion of input data from one or few
tasks can be utilized in first step. The major challenge in catastrophic forgetting is the performance on a previously
learned task can not be significantly degrade
over time when sequentially recent tasks are included in the learning process. In other words, the learner must identify how to classify within each task, as it is considered into test samples at inference time \cite{boschini2022class}. This is a direct result of a more general problem in neural networks, namely the stability-plasticity dilemma as discussed in previous paragraph. To sum up,  plasticity implies the ability of adaptation recent outcomes, and stability considers the usage of former outcomes \cite{de2021continual}. 

Continual learning was initially conducted for computer vision tasks. For that reason, the studies dealing with computer vision tasks to continual learning include the concatenation of multiple datasets to track data drifts appeared in data distribution especially on large-scale streams \cite{cossu2022class}. The learning objective was the classification problem of patterns by assuming that which dataset was used to gather the pattern. In other words, tasks are labelled to facilitate the continual learning problem, since patterns from different datasets can be easily eliminated by the model during both training and inference. To summarize, if task incremental is only considered,
where data was collected sequentially as batches and one batch corresponds to one task, a recent set of labels
were prepared to be trained. In other words, we assume for a given task, all data  is considered for offline training iteratively. This provides learning from multiple epochs over all its training data. 
On the other hand, data from former or further tasks is not reachable. Optimizing for a new training task in this setting will cause catastrophic forgetting, with significant drops in performance for old tasks \cite{parisi2019continual}

Besides task incremental based studies, the attention of the community has now focused to class-incremental continual learning \cite{kueffner2023into}, where experiments are conducted on a single dataset, and the dataset was splited considering class without any previous information about the task label both for during training and during inference. This implies that the task identifier have to be determined together with sub-class label. The main challenge, which class-incremental methods have
mentioned, is balancing the different classifiers. The
imbalance problem arouses since the training phase of the
recent task, there is none or only limited data was gathered
from former tasks, which biases the classifier towards the
most recently learned task\cite{liu2020generative}. To overcome this issue, solutions including replay scenarios have been conducted. iCarL \cite{rebuffi2017icarl} states a limited number of exemplars from former tasks in such a way that exemplars converge the mean of classes in the feature space. \cite{wu2019large}
stated that the last fully-connected layer tends to emphasize on new classes, and  a linear model estimated from exemplars was proposed for this issue. \cite{hou2019learning} used the cosine similarity-based loss, which, combined with exemplars. These overall methods need to use storage of exemplars. On the other hand, especially privacy concerns or storage restrictions force not to store any exemplars from previous tasks.

Class incremental learning more general and regarded as more
realistic than the other scenarios in continual learning \cite{farquhar2018towards}. Methods capable of working in a single-head assumption are usually divided into two categories:regularization-based and rehearsal-based.The regularization terms ensure
to lead the optimization towards a good balance between stability and plasticity. Specifically, a penalty term enable to control the changes in the weights for the former tasks while the new tasks are being learned \cite{kirkpatrick2017overcoming}. These methods can be effective for midsize ordered
tasks but generally usefulness to scale these methods to complex problems is controversial. On the other hand, rehearsal models use the
memory buffer to store similar elements from former
observations. iCaRL , ER , SER , TEM , CoPE can be given as the main examples for the rehersal methods \cite{de2021continual}. Rehearsal methods explicitly retrain on
a limited subset of stored samples while training on new
tasks. The performance of these methods is upper bounded
by joint training on previous and current tasks. Most notable
is class is iCaRL as discussed in previous paragraph. 
Experience Replay (ER) (\cite{robins1995catastrophic} solely replays the stored items together with the input stream as the joint training to replicate the training as an independent and identically distributed task.In addition, this method has been stated to be highly effective even with a minimal memory footprint in terms of the selection of samples to obtain in the memory buffer or the sampling of examples from it .For data incremental learning, \cite{rolnick2019experience} states reservoir sampling to set bounds to the number of stored samples to a fixed budget assuming the data stream. Continual Prototype Evolution (CoPE) \cite{de2021continual2} merges
the nearest-mean classifier solution with an efficient
reservoir-based sampling task. Additional experiments
on rehearsal methods to enhance class incremental learning are still being studied. Note that rehearsal methods tend to overfitting issue and naturally, the subset
of stored samples and limited as joint training,
constrained optimization is an alternative solution for backward/forward transfer. As proposed in GEM \cite{lopez2017gradient},the main approach is to only constrain new task learning not to forget with former tasks by monitoring the estimated gradient direction on the feasible region outlined
by former gradients. 

\textbf{Motivation}. 
Deep neural networks exhibit extraordinary performance across a myriad of learning tasks. However, the methodologies employed in the domain of continual learning have not yielded promising outcomes. Nevertheless, the commendable aspect lies in the discernment of the problem's definition and the determination that it is amenable to resolution. This study attempts at translating the theoretical underpinnings of Martin Heidegger's work (being and time)( \cite{heidegger2010being}) into the classical deep learning paradigm with the objective of addressing the challenges inherent in continual learning.

In his most influential book called being and time, Heidegger says: "When Dasein directs itself towards something and grasps it, it does not somehow first get out of an inner sphere in which it has been proximally encapsulated, but its primary kind of Being is such that it is always 'outside' alongside entities which it encounters and which belong to a world already discovered." In order to avoid from a philosophical discussion, we consider Dasein to be synonymous with the human being. And we try to implement the aforementioned learning style of humans into deep learning paradigm. 
Going back to the quote, we have noticed that in the classical deep learning paradigm, the model actually does get out of its inner sphere and this may be the leading cause of the forgetting. For this reason, we attempted to limit the movement of the model by setting up a negotiation table between the model and the labels that correspond to the samples. We investigated the limits of the movement restriction. Because we know that if we limit the movement so much, the model won’t be able to learn the task. Also, if the limits are loose, the model will forget the task so quickly. 

\begin{figure}[ht]
\centering
\includegraphics[width=0.999\textwidth]{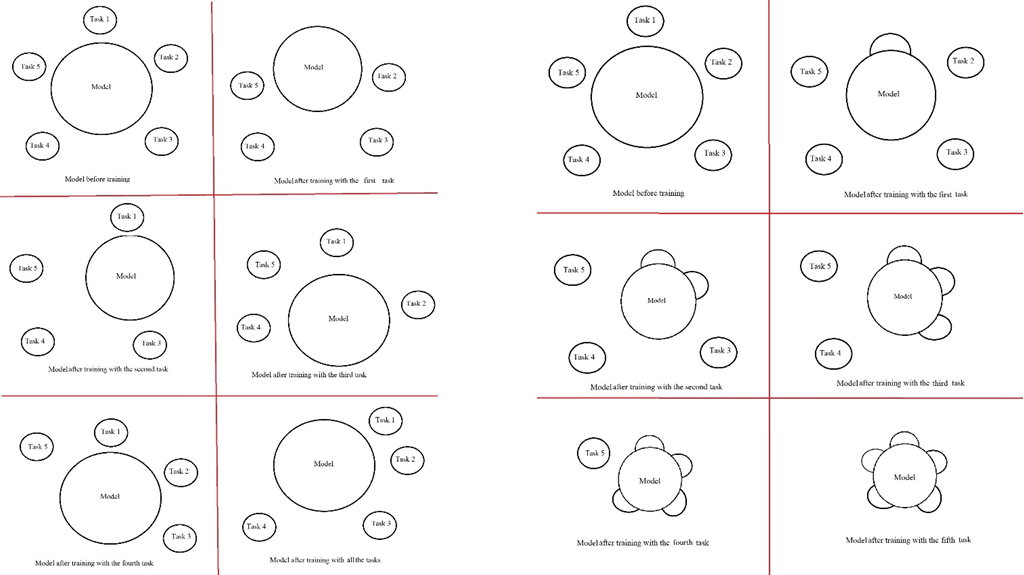}
\caption{Classical vs Negotiated training comparison. The purpose of negotiation is to prevent the model from completely leaving its neighborhood when learning new tasks. Because if the model has complete freedom to move, forgetting the previous tasks is inevitable. }
\label{fig:classical_vs_negotiated}
\end{figure}

A visual representation of the paradigm is presented in Figure~\ref{fig:classical_vs_negotiated}. On the left side of the figure, the classical training paradigm is depicted. It demonstrates how forgetting may be related to the movement of the sphere that is associated with the model weights. 
On the right side of the figure, the negotiated training paradigm is depicted. It demonstrates how forgetting may be prevented by restricting the movement of the model.

\textbf{Findings.}

The obtained results of our experiments are highly favorable. As illustrated in Table 1, the average classification accuracy has notably increased. To underscore the non-random nature of this improvement, we conducted experiments with five different seeds and presented their averages in the table. For further details, readers are encouraged to execute the code on their own.

These findings were derived from codes written in a rudimentary manner, as our primary objective was not to achieve exceptional accuracy. Rather, our intent is to demonstrate that the method we advocate is noteworthy and warrants a series of subsequent investigations. To facilitate reproducibility, we have shared the codes on GitHub. Unfortunately, we encountered difficulties in executing other relevant scripts found in the literature, attributing this challenge partially to our own limitations and partly to potential idiosyncrasies in those scripts. The notebooks we provided are designed to be executed directly, yielding conclusive results.

Furthermore, as evident in Figures 6, 7, 8, and 9, we observed distinct variations in the initial negotiation rate for each dataset, contingent upon the starting point. We posit that an increase in the number of model parameters necessitates a corresponding increase in the initial negotiation rate. However, it is worth noting that model parameters somewhat depend on the number of classes and other properties of the dataset at hand. The intricacies of this relationship remain not entirely clear to us; nevertheless, it appears to be of substantive significance. We welcome further clarification from the continual learning community.

\textbf{Paper overview.}
This paper focuses on the problem of catastrophic forgetting issue that aroused from supervised deep learning tasks as a part of class incremental learning and aims to overcome this issue by using negotiated representations of learning process. This paper conducts the balance between previously learned information and recent extractions from iterations and enhance the preservation of necessary information coming from each iteration. This paper is organized as follows: i)we gave a brief literature review for continual AI, catastrophic forgetting and class incremental learning .ii) we discussed the proposed method's fundamentals such as divergence Based Feature Extractor,Walsh Matrix and negotiated training. iii) we also refer the difference of our approach that includes training paradigm and model architecture. 

\section{Literature Review and Background }
Catastrophic forgetting is a complex phenomena involving many practical
considerations. \cite{ratcliff1990connectionist} presents an extended demonstration and exploration of catastrophic
forgetting in a back-propagation network. \cite{ratcliff1990connectionist} simple demonstration of
this forgetting effect emphasis on to preserve necessary information to the end of simulation or iteration. 

In this section, we gave fundamental information for Continual Learning and negotiated training. Continual learning (CL) aims to design algorithms that learn from non-stationary sequences of tasks which are being able to reapply, and adapt previously learned abilities to new situations \cite{khetarpal2022towards}.  

Classically,the main challenge of CL is catastrophic forgetting (CF) – fast performance degradation on previous tasks when learning from new data. CF is usually evaluated on scenarios with sequences of disjoint tasks \cite{lesort2022challenging}. Another issue for CF is that no one can guarantee  what sort of data the network is utilizing to learn.This means the data can vary overtime and naturally, the outputs can also vary. If the model tend to be trained with data that is too different from its fundamental training, it may experience catastrophic forgetting.In order to handle catastrophic forgetting issue, machine learning systems must have the ability to gather recent knowledge and refine existing knowledge considering continuous input and, prevent the novel input from significantly interfering with existing knowledge. The ability of a system must be plastic to include recent  information and robust not to miss the consolidated knowledge is named as the stability–plasticity dilemma and has been widely studied in computational models \cite{ditzler2015learning,mermillod2013stability}. 

The literature is mainly based on the mitigation of catastrophic forgetting that utilizes memory systems that store previous data and regularly replay old samples interleaved with samples drawn from the new data \cite{robins1995catastrophic}, and these methods are still used today. However, a general drawback of memory-based systems is that they need the considerable amount of data storage of former information, leading to large working memory requirements. Thus,  protection of consolidated knowledge
from being overwritten by the learning of recent information can be more reliable to overcome memory requirements. \cite{parisi2019continual}. 

This study emphasizes on class incremental learning, where the goal is to extract a unified classifier over incrementally occurring sets of classes. Because of all the incremental data cannot be evaluated for unified training, the major challenge is to avoid forgetting former classes while learning new ones. The three vital issues of a class incremental learning algorithm
obtain the memory buffer to store few exemplars from old
classes, a forgetting constraint to keep formerly learned knowledge
as learning from new tasks, and a learning mechanism that prevent imbalance problem from old and new classes. Although numerous studies have been emphasized to overcome each of these issues, general understanding of best practices is still a gap for researchers \cite{mittal2021essentials}. 

\subsection{Class incremental continual learning}

The objective of class-incremental learning (class-IL) is to learn a unified classifier from a sequence of data from different classes.Data arrives incrementally as a batch of
per-class sets ${X}$ in such a way that $\{X^0, X^1, \ldots, X^{t}\}$ where ${X^y}$
contains all images from class y. Learning from a batch of classes
can be considered as a task $T$, At each incremental step,
the data for the new task $T_i$ arrives, which contains samples
of the new set of classes. At each step, complete data is only available for new classes $\{X^s+1,  \ldots, X^{t}\}$. Only a small amount of exemplar data from previous classes $({X^1,  \ldots, X^{s}})$ is retained in a memory buffer of limited size. The model is expected to classify all the classes seen so far \cite{mittal2021essentials}.

In other words, Class-Incremental Continual Learning (class-IL), a model $ f(\cdot, \theta) $ is trained on a sequence of $T$ tasks $\{T_0, T_1, \ldots, T_{t-1}\}$ having access to one at a time. The $i^{th}$ task consists of datapoints $\left(x_i^{n}, y_i^{n}\right)_{(n=1)} ^ {[T^i]}$ where $y_i^{n} \in Y_i^{n}$ with disjoint ground-truth values for different tasks, i.e.,  $y_i \cap y_j = \emptyset$ such that $i \neq j$. For the sake of simplicity, we assume all tasks have the same number of classes, i.e., $| Y_i | $ = $| Y_j |$ = $| Y | $. While all ${x_i}$ are $i.i.d. $ within $T_i$ the overall training procedure does not abide by the $i.i.d. $ assumption, as input distribution shifts
between tasks and labels change. The objective of class-IL is the minimization of the risk over all tasks as $Loss$ = $\sum_{i=0}^{T-1} E_{(x,y) \rightarrow \mathcal{T}_i} \left| L(f(x;\theta),y \right|$
where L stands for the loss (e.g., the categorical crossentropy) of predicting $f(x,\theta)$ given $y$ as the true label. The optimal solution  $\theta^ * $ should provide accurate predictions for
all tasks; however, this has to be pursued by observing one
task at a time. To account for this, its actual learning objective should combine the empirical risk on the current task $T_c$ with a separate regularization term $L_R$ as:

\[
\text{Loss} = E_{(x,y)\rightarrow T_c} \left(|L(f(x;\theta),y)| + L_R\right)
\] 

The second term $L_R$ serves a twofold purpose: i) it prevents
the model from forgetting past knowledge while fitting new
data \cite{robins1995catastrophic} ; ii) it encourages the learner to gather per-task classifiers into a single and harmonized one \cite{boschini2022class}.

Class-IL models are evaluated using three metrics: average incremental accuracy, forgetting rate and feature retention. After each incremental step, all classes seen so far are evaluated using the latest model. After $N$ incremental tasks, the accuracy $A_n$ over all $(N + 1)$ steps is averaged and reported. It is termed as average incremental accuracy (Avg Acc), introduced by \cite{rebuffi2017icarl}.Also, the evaluation of the forgetting rate $F$ proposed by \cite{liu2020generative} is crucial. The forgetting rate measures the performance drop on the first task. It is the accuracy difference on the classes of the first task $\{X^1,X^2, \ldots, X^s\}$ for the test set using $\theta^ * $. Therefore, it is independent
of the absolute performance on the initial task $T_0$. In addition to these metrics, retention in the feature extractor measures how much information is retained in the feature extractor while learning the
tasks incrementally as compared to a jointly trained model \cite{mittal2021essentials}.

\subsection{Drawbacks of class incremental learning }

The fundamental obstacles to effective class-incremental
learning are conceptually simple, but in practice very challenging to overcome. These challenges originate from the sequential training of tasks and the requirement that at any
moment the learner must be able to classify all classes from
all previously learned tasks. Incremental learning methods must balance retaining knowledge from previous tasks
while learning new knowledge for the current task. This
problem is called the stability-plasticity dilemma. A naive
approach to class-IL which focuses solely on learning the
new task will suffer from catastrophic forgetting: a drastic
drop in the performance on previous tasks. Preventing catastrophic forgetting leads to a second important
problem of class-IL, that of intransigence: the resistance to
learn new tasks . There are several causes of catastrophic
forgetting in class-incremental learners:

 Weight drift: While learning new tasks, the network
weights relevant to old tasks are updated to minimize
a loss on the new task. As a result, performance on
previous tasks suffers – often dramatically. 
Activation drift: Closely related to weight drift, changing weights result in changes to activations, and consequently in changes to the network output. Focusing on
activations rather than on weights can be less restrictive
since this allows weights to change as long as they
result in minimal changes in layer activations. 
Inter-task confusion: in class-IL, the objective is to
discriminate all classes from all tasks. However, since
classes are never jointly trained, the network weights
cannot optimally discriminate all classes. 
This holds for all layers in the network. 
Task-recency bias: Separately learned tasks might have
incomparable classifier outputs. Typically, the most
dominant task bias is towards more recent task classes.
This effect is clearly observed in confusion matrices
which illustrate the tendency to correctly classify inputs as
belonging to the most recently seen task.

The first two sources of forgetting are related to network
drift and have been broadly considered in the task-IL literature. Regularization-based methods either focus on preventing the drift of important weights or
the drift of activations. The last two points are specific to class-IL since they
have no access to a task-ID at inference time. Most research
has focused on reducing task imbalance,
which addresses the task-recency bias. To prevent intertask confusion and learn representations which are optimal
to discriminate between all classes, rehearsal or
pseudo-rehearsal are commonly used.

To overcome these issues, negotiated training is proposed. In this concept, negotiation mechanism enable not to forget previous tasks outcomes by not allowing completely leaving its neighborhood before a new learning task initials. This perspective has experienced on Split MNIST, Split Fashion MNIST, Split CIFAR 10 and Split CIFAR 100. The results are also discussed in the relevant section.

\section{The Proposed Method}

\begin{figure}[ht]
\centering
\includegraphics[width=0.9\textwidth]{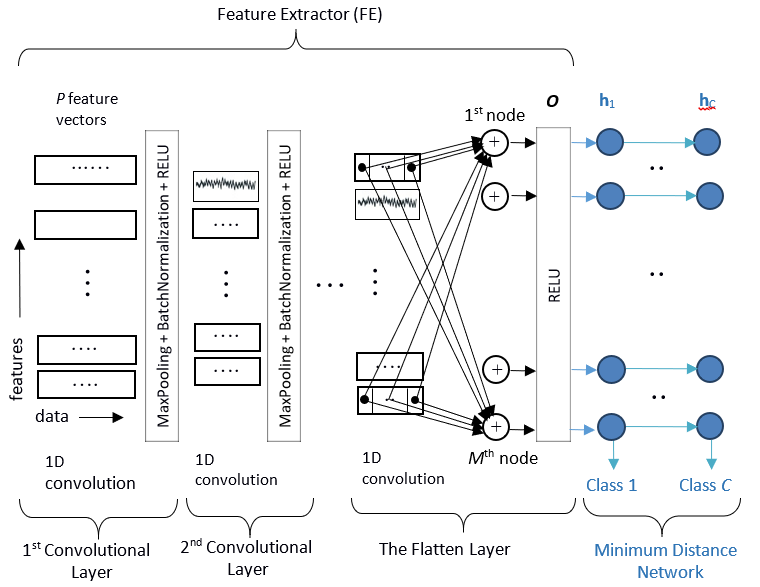}
\caption{Model Architecture}
\label{fig:Main_Model}
\end{figure}

The models that are tailored here are each a variant of the main model that is proposed for the classification of EEG/ECG signals in the study \cite{olmez2021strengthening} and depicted in Figure ~\ref{fig:Main_Model}. The main difference of the proposed method is negotiation paradigm that is diversified from the classical machine learning perspective. In traditional experiments, an isolated way is generally conducted in such a way that model is learning from the tasks separately and can not inherit from other task related learning . For instance, suppose that we have 5 tasks as illustrated in Figure~\ref{fig:classical_vs_negotiated} and the model should learn from these 5 different tasks. In the first phase the training is executed from Task 1 and the other four tasks are ignored. After that, model learns from Task 2 but forgets the first task related outcomes. Consequently, model solely learns from the relevant task and ignores the outcomes extracted from other task related trainings.  This process is ineffective for every machine learning based cases. To overcome this ineffectiveness, this paper proposed an iterative perspective for different tasks that does not miss the relevant information coming from former tasks' training. For example, Task 1 is executed and relevant outcome is stored before the implementation of Task 2. After Task 1 and Task 2 are executed, then Task 3 remembers the outcomes from Task 1 and Task 2 and then modify itself according to these results. The other consequent phases are obvious that negotiated learning considers other task related trainings iteratively and this perspective does not miss task related information that the trainings end in a considerable time and epoch with a reasonable performance.

\subsection{Critique of the One-hot representation in continual learning}

The issue with one-hot encoding arises when the model is trained on each class pair independently, without considering previous learned tasks or possible future tasks. Specifically, if we consider a binary classification scenario, the model aims to activate either Neuron 1 or Neuron 2, while simultaneously deactivating all other neurons. This method forces the model to optimize based on just the two classes under consideration, ignoring any information that may be relevant for determination of other classes.
In a continual learning framework, this approach is problematic because it prevents the model from effectively leveraging its past learning experiences when training on new tasks. In essence, the model starts learning each new task from scratch, resulting in a high risk of catastrophic forgetting - where knowledge of previously learned tasks is lost. This is a crucial issue in continual learning models that prevent to learn new tasks while preserving performance on prior tasks.

In other words, conventional one-hot encoding based models can fail to effectively balance between preserving learned knowledge from prior tasks and accommodating new knowledge from current tasks. This can potentially compromise the model's ability to learn continually and adapt to new tasks over time. Hence, it is critical to devise more sophisticated class representation and learning strategies for continual learning scenarios to overcome this limitation.

\subsection{Divergence Based Feature Extractor}

\begin{figure}[ht]
\centering
\includegraphics[width=0.50\textwidth]{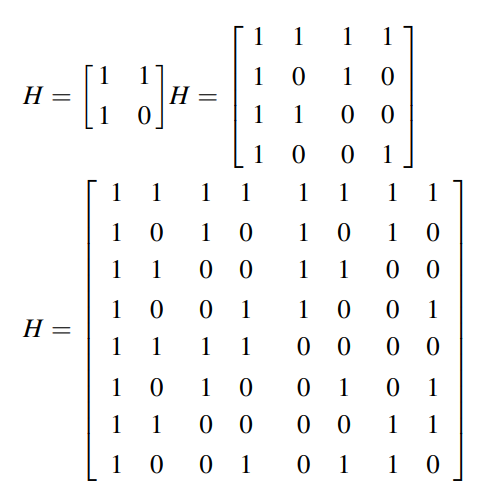}
\caption{Creation of Walsh matrix: Depending of the size of the vector representations we need, we can always create a bigger walsh matrix to select representations from. However, once we run out of vectors in the walsh matrix, we cannot generate a new walsh vector. Therefore, the walsh vector size should be selected big enough to ensure we won't run out of vectors. }
\label{fig:walsh_Vectors}
\end{figure}

In this study, a divergence-based feature extraction technique is applied within the domain of continual learning. This approach leverages vector representations that maintain a constant Hamming distance between each pair of classes, facilitating robust class separability.
To illustrate, consider four classes A, B, C, and D, represented by binary vectors as follows:
Class A: 11111111
Class B: 10101010
Class C: 11001100
Class D: 10011001
(see Figure~\ref{fig:walsh_Vectors})

\begin{figure}[ht]
\centering
\includegraphics[width=0.5\textwidth]{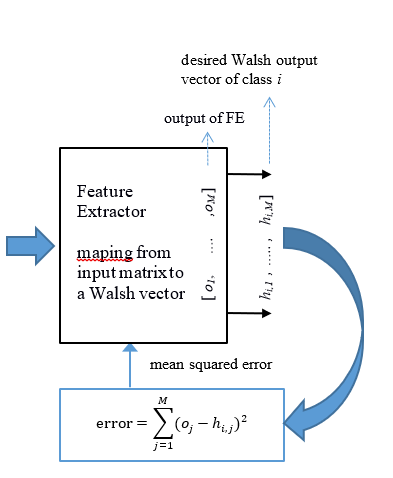}
\caption{MDN Classifier and Loss Calculation}
\label{fig:mdn_classifier_and_loss_calc}
\end{figure}

Upon training the model with the first pair of classes (A and B), it learns not only the mutual features shared by these two classes (neurons 1,3,5,7), but also the distinctive features that uniquely belong to each class(neurons 2,4,6,8). When it is time to train the model with the next class pair (C and D), we do not directly continue to the learning process. Instead, we generate predictions for all possible outcomes of the model and calculate the loss relative to all available representations. The representation with the minimum loss is then assigned as a placeholder for the corresponding class label.
In the classification of the samples, MDN (Minimum distance Classifier) is utilized (see Figure~\ref{fig:mdn_classifier_and_loss_calc}. This method categorizes feature vectors by determining the shortest distance between an input vector and all class centers. The vector is then assigned to the class center that exhibits the shortest distance. This approach is straightforward and effective for solving classification problems ~\cite{mdn_classifier}.

MDN employed to categorize unknown image data into classes by minimizing the distance between the image data and the class in a multi-feature space. The distance, serving as a measure of similarity, is defined in such a way that the minimum distance corresponds to the maximum similarity. It should be kept in mind that this classifier can be replaced with other classifiers. For instance, SVM could also be a better option. On the other hand, MDN does not require any additional training for calculating the similarities. 

The primary advantage of using divergence-based vector representations in this continual learning context relies on the selection of the closest vector representation when introducing a new class pair. This approach minimizes the amount of training required for the model, thereby reducing the potential for catastrophic forgetting.

Moreover, this divergence-based approach makes use of mutual and distinctive information across classes. It allows the model to learn shared features that generalize well across classes while also capturing unique characteristics that differentiate each class. This results in more robust models that can handle new learning tasks with minimal impact on the performance of previously learned tasks.

\subsection{Negotiated Training}

The training paradigm is first proposed in
\cite{korhan2023negotiated} to deal with overfitting problem. The initial intention of negotiated training was to deal with uncertainties of the labeled data. That is, no observation from any perspective should be assigned to any class with absolute certainty. The example give is that as the quality of observation decreases the probability of all the possible classes should approximate to each other. Since there is no reliable way to assign the probabilities, the authors decided to solve this problem by setting up a negotiation between the provided labels and model's interpretation.

However, in this study, we used negotiation paradigm for the gradual allocation of the model's capacity to new tasks. That way, the model is exploited in a unique and successful way to achieve high accuracy in the new tasks while still remembering the other tasks. 

A numerical example can be given for this issue: Let's assume that negotiation rate is 0.8. 
The first task will use 20 percent of the model's capacity. Note that the model still has 80 percent capacity to use. 
Then, the second task will take 16 percent of the total capacity from the model and 4 percent of the total capacity from the previous task. Now the model has 64 percent capacity to use. Notice that the model begins to forget the first task already. but after 5 tasks the model still has 8 percent of its capacity allocated for the first task. And this amount is enough to remember some aspects of the first dataset.

\begin{figure}[ht]
\centering
\includegraphics[width=0.99\textwidth]{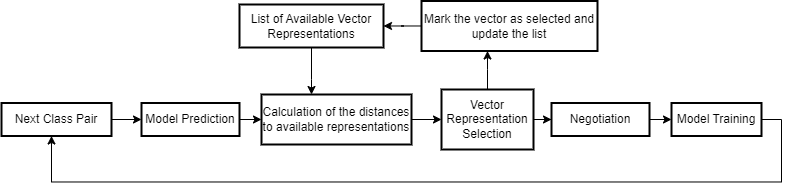}
\caption{Training Paradigm of the model}
\label{fig:Training_paradigm_big}
\end{figure}

Training paradigm is summarized in Algorithm 1.
Also, schematic illustration of the process is given in Figure ~\ref{fig:Training_paradigm_big}

\begin{algorithm}
\caption{Continual Learning with Vector Representations}
\begin{algorithmic}[1]
\Require Vector Representation Size
\Require Negotiation Rate
\Require Negotiation Plasticity Rate
\Require Number of Training Epochs Steps for Continual Learning
\For {n in Number of Tasks}
    \State Current Task $\gets$ Dataset(n)
    \For {m in Number of Classes in the Task}
        \State Predict the Output Representations
        \State Find Nearest Vector Representation
        \State Allow the model to negotiate with assigned representations for each sample
        \State Calculate the negotiation rate for the next task. So that model's capacity is allocated equally to each task
    \EndFor
    \State Train the Model

    \For {k in Trained Tasks}
        \State Calculate Test Accuracy of the Model for all tasks the model has seen so far. 
        \State Preferably print the test accuracies for demonstration purposes. 
    \EndFor
\EndFor
\end{algorithmic}
\end{algorithm}

\subsection{Gradual Allocation of the Model's Capacity for Training New Tasks}

We have demonstrated that negotiation paradigm allows the model to remember the previous tasks.
But the problem was that  although this paradigm allows the model to remember the previous task, it always has the most allocated capacity for the last task. In order to allocate approximately same amounts of capacity of the model for each task, we have developed the negotiation rate scheduler. It resembles to the learning rate scheduler in the literature. However, for continual learning perspective, it is much better at protecting the previous tasks.  The reason for this phenomenon is that the learning rate does not prevent the training mechanism from eroding the the previously learned tasks. Especially if we increase the number of iterations or we have a large dataset, learning rate scheduler will inevitably fail to do its job. The model learns from the classes against each other, and the contrast between any two class erodes everything that is not present. negotiation rate scheduler enables to allocate a significant part of the model's memory , which was initially not very effective, for new tasks.

\color{black}

\subsection{Verification of model allocates equal amount of its capacity for each task }

The problem can be numerically explained as follows: Let's assume that the negotiation rate is 0.9. The model 10 percent of its capacity to learn the first task. Then the model goes on to be trained on the second task with the same negotiation rate. It again uses 10 percent capacity of the model's capacity. However, that 10 percent also includes the part that the first task was trained. Therefore, the model's capacity that is allocated to learn the first task becomes 9 percent overall while it is 10 percent for the second task. 
In order to allocate equal amount of model's capacity to each task we have developed a formula given below:
 
\begin{equation}
    \text{optimal negotiation plasticity} = \frac{1}{{2 \cdot \text{neg} - \text{neg}^2}}
\end{equation}

The optimal negotiation plasticity above provides the model's allocated capacity remains the same for all consecutive tasks. 
 \cite{hadsell2020embracing}. Let's consider an illustrative example with a negotiation rate (denoted as $n$) initially set to 0.8. The model allocates 20 percent of its capacity to the first task. Surprisingly, if we maintain this negotiation rate, the model also assigns 20 percent capacity to the second task. However, the capacity allocated to the first task decreases to 16 percent due to the compound effect — 20 percent of 20 percent results in a reduction of 4 percent. To ensure equal capacity allocation for each task, we aim to update the negotiation rate ($n$).

Let's assume we want the model to allocate 10 percent capacity to the first task ($n = 0.1$). Since we intend to teach the second task as well, this capacity needs to be scaled by a factor ($m$). We desire this scaling to be such that the capacity assigned to the second task ($1 - m$) equals $1 - n$. This leads to a mathematical relationship $(1 - n) \cdot m = 1 - m$, and solving this equation yields $m = \frac{1}{{2 + n}}$. Here, $m$ represents the new negotiation rate. Importantly, we introduce a new variable, denoted as $\frac{m}{n}$, which characterizes the factor by which $n$ should be multiplied to obtain the new negotiation rate $m$. This mathematical exploration helps us find an optimal adjustment for the negotiation rate that ensures balanced capacity distribution across tasks.

\subsection{Proposed model} 

The proposed model is configured based on a supervised manner by using the training samples and their corresponding vector representations that are previously negotiated with the model. As seen from Figure 4, the model has convolutional layers with a specified number of neurons. Relu was selected as the activation function for hidden layers and Max-pooling2D is adapted with fully connected layers.Output layer uses the  size of Walsh matrix.  Note that the last fully connected layer has N output neurons since each class will be assigned to a vector of the length N. 

Another contribution of the proposed model is defining a custom sigmoid function which turns sigmoid function into a function by which 
class memberships are represented more linearly.The need to use the modified sigmoid function is that if logistic functions used, even the slightest change cause a dramatically change as seen in the output.This situation directly affects model training efficiency. The softened sigmoid function is given in the following equation:

\begin{equation}
f(x) =
\begin{cases}
    \frac{1}{{1 + \exp\left(-\frac{x}{{\text{{sigmoid softening coeff}}}}\right)}}, & \text{if sigmoid softener = True} \\
    \frac{1}{{1 + \exp(-x)}}, & \text{otherwise}
\end{cases}
\end{equation}
The other point for the proposed approach is the appearance of Walsh matrix. This function determines a matrix whose rows are the vectors that can be used to represent classes. Note that Walsh vector representations allow classes to share some features for class incremental learning. In other words, Walsh representations allow previous tasks dynamically create features that can be exploited by future tasks.This point of view increase the efficiency of model training [URL-1]. 

The flow of the proposed approach can be given in the following: 

1-) Create a Walsh matrix that contains N number of Walsh vectors: To dynamically create features that can be exploited by future tasks, Walsh matrix is used for vector representations.The size of Walsh matrix is determined as  \begin{align*}
    \text{{size\_of\_walsh}} &= \lfloor \log_2(\text{{size\_of\_walsh}}) + 1 \rfloor
\end{align*}

2-) Split the training and test samples of each class pair: To measure the performance of the proposed model, test and train split for the dataset is mandatory.

3-) Select all the samples of each class of the class pair that the model will be trained with: The samples are collected for training phase for each class pair.

4-) Predict the outputs of the samples of the selected class:In class incremental learning, when  a new class is gathered, a class representative vector is assigned to that class. A class representative vector must be selected among the vectors that are not used. For this purpose, we define a list called 'available representations' to make sure that we don't use any of the vectors that are previously used for representing another class. 
    
In order to better exploit the model, we first calculate output representations of all samples. 
We then calculate the distances of all outputs to each available representation and take the mean of the distances, so that  it can be comparable for all the mean values. This procedure facilitates the decision of which representation is the closest candidate, which implies the best, for representing the new class.

5-) Select the nearest Walsh vector as representative of the corresponding class: We select the closest vector as a class representative update by changing the index 'available representations' as "FALSE".  By updating that index, the selected index is ensured that won't be used as the class representative for another class anymore.

6-) Set up a negotiation between the output representations and model's predictions: Finally, output representations negotiate with the past experiences in order to not forget previous tasks. Here, optimal plasticity for the allocation of equal amount of model's capacity for all the tasks is considered. 

Here, selection of the representative vectors and negotiation phase for each class pair takes place after the model is trained with the previous class pair.

For the evaluation phase,the predicted results are compared with the Walsh vectors.After output representation
for one sample was obtained, the distance between the obtained representation and previously assigned class representations was calculated based on 'Binary Cross Entropy loss- BCE'. If a new class is introduced into the model, vector representation was given to that class must be tracked as determined Walsh vectors as output labels.This distance can be represented the mean of loss function with an epsilon value as follows:

\begin{equation}
\begin{aligned}
    \text{loss\_calc} = - \Bigg(&y_{\text{train\_i\_predicted\_mean}} \cdot \log\left(\text{custom\_sigmoid}(\text{walsh\_vec} + \epsilon)\right) \\
    &+ (1 -y_ {\text{train\_i\_predicted\_mean}}) \cdot \log\left(1 - \text{custom\_sigmoid}(\text{walsh\_vec} + \epsilon)\right)\Bigg)
\end{aligned}
\end{equation}

In class incremental learning, if a new class is introduced, then a class representative vector to that class must be determined. For further progress, a class representative vector among the vectors that are not used must be selected.  For this purpose, available representations list is defined considering that any of the vectors that are previously used for representing another class is not reused in advance.  The step by step explanation of this procedure for each unique class in output train set is as follows: 

1:Predict the mean output for the class using the model.

2:Find the closest available Walsh vector to the predicted mean.

3:Update the assignment matrix with the chosen vector for the class.

4:Mark the chosen vector as not available for future assignments.

5:Return the updated assignment matrix and check the availability status of each vector.

\section{Experiments}

\subsection{Dataset characteristics}

The experiments for the negotiation approach is conducted on  Split MNIST, For each dataset, Split Fashion MNIST, CIFAR 10 and CIFAR 100. For each task, there are 5 consecutive tasks that contain the samples from certain class labels.The first dataset is Split MNIST that serves as a benchmark dataset in research related to continual learning, incremental learning, and lifelong learning. It is important to note that the construction of the Split MNIST dataset can vary, and different studies may adopt slightly different configurations. The specific assignment of labels to tasks might differ, allowing researchers to create custom splits tailored to the requirements of their experiments.

The Split MNIST dataset is a modified version of the widely used MNIST dataset, a frequently employed resource as a benchmark dataset in machine learning. MNIST comprises 28x28 pixel grayscale images featuring handwritten digits from 0 to 9, each associated with a corresponding label.
In the context of Split MNIST, the dataset undergoes segmentation into multiple tasks, each linked to a subset of digit labels. This partitioning aims to replicate a continual or incremental learning environment, where a model systematically acquires skills for different tasks.For instance, in a Split MNIST scenario involving two tasks:
i)Task 1: Encompasses images of digits 0 and 1.
ii)Task 2: Encompasses images of digits 2 and 3 and so forth.

In other words, for Split MNIST, the initial task encompasses all samples corresponding to labels 0 and 1, while the subsequent task comprises samples associated with labels 2 and 3, and so forth.The objective is to assess the model's capability to learn new tasks while retaining knowledge from previous ones. This setup proves valuable in situations where a model must adapt to fresh information continuously, avoiding the need for complete retraining on the entire dataset.

The second benchmark dataset is Fashion MNIST dataset that was designed as a drop-in replacement for the original MNIST dataset and contains a collection of grayscale images of 10 different fashion categories such as shirt, sandal, dress etc. Each image is a 28x28 pixel square, similar to the original MNIST dataset.
The reasons to choose Fashion MNIST is that provides a more complex and diverse set of categories compared to the original MNIST dataset. This diversity makes it suitable for training models that need to recognize and adapt to different types of objects, which is a common requirement in continual learning scenarios.The images in Fashion MNIST are also more complex than the simple handwritten digits in MNIST. This increased complexity allows models trained on Fashion MNIST to learn more intricate features, making them potentially more adaptable to a wide range of visual tasks.As Fashion MNIST has a similar structure to the original MNIST, it allows for easy comparison with models trained on the classic dataset as a scalability of the approaches.

The CIFAR-10 dataset is a widely used computer vision dataset that consists of 60,000 32x32 color images in 10 different classes, with 6,000 images per class. The dataset is split into 50,000 training images and 10,000 test images. Each class represents a specific object or category such as dog, cat, frog, horse etc.The main advantage of CIFAR-10 is that covers a diverse set of object categories, making it suitable for training models on a broad range of visual concepts. This diversity allows models to learn a rich set of features and representations.Note that a model is trained on a sequence of tasks over time. Each class in CIFAR-10 can be considered a task, and training on this dataset can help evaluate a model's ability to adapt to new tasks without forgetting previously learned information.

CIFAR-100 consists of 60,000 32x32 color images in 100 different classes. Each class contains 600 images, with 500 for training and 100 for testing.The dataset is organized into a hierarchical structure with 20 superclasses, each containing 5 fine-grained classes.CIFAR-100 is more challenging than its counterpart CIFAR-10, as it involves a larger number of classes and a finer level of classification.

For Continual learning, CIFAR-100's hierarchical structure and diverse set of classes make it suitable for experiments. Note that models need to adapt to new tasks without forgetting previously learned ones in continual learning and it is critical  to evaluate models with these numerous number of classes for catastrophic forgetting where performance on early tasks degrades when learning new tasks.

\subsection{Model Parameters}

Optimizing model parameters is a cornerstone in the field of continual learning, as it enables models to effectively retain and apply knowledge from previous tasks when addressing new ones. Parameters serve as a reservoir of the model's accumulated experiences and are crucial for maintaining a balance between absorbing new information and preserving essential insights from past data. The careful adjustment of these parameters is central to our strategy, ensuring a nuanced blend of stability and plasticity in the model's learning process. Such management is vital to prevent catastrophic forgetting, where a model may lose the ability to recall what it has learned previously.

Our experimental framework is designed to rigorously test the efficacy of the proposed approach, utilizing specifically configured models for each dataset to best demonstrate their learning capabilities. For a comprehensive overview of the experimental configurations and additional insights, please refer to our detailed documentation available on GitHub ~\cite{github_page}.

\subsubsection{Model Configurations for Datasets}
Each dataset in our study required a different model to optimize the balance between parameter retention and adaptability for new information:

\begin{itemize}
    \item \textbf{Split MNIST:} The model consists of 2 convolutional layers, 2 max-pooling layers, 1 flattening layer, and 1 dense layer, with a total of 44,432 parameters. This architecture was chosen for its simplicity and efficiency, providing a strong baseline to evaluate the performance on basic image classification tasks.
    \item \textbf{Split Fashion MNIST:} We employed a model with a sequential structure including one convolutional layer and max-pooling layers, concluding with a dense layer. This model has 692,928 parameters, a design choice driven by the need to capture more complex patterns within the dataset while maintaining computational feasibility.
    \item \textbf{CIFAR-10:} The model for CIFAR-10 is a sequential structure with six convolutional layers and 3 max-pooling layers, ending with a dense layer, comprising 583,502 parameters. This configuration is suitable for the dataset's broader range of features, requiring a better feature detection capacity.
    \item \textbf{CIFAR-100:} For the CIFAR-100 dataset, with its substantial complexity and class variety, our model includes 6 convolutional layers, 3 max-pooling layers, and a dense layer after flattening, totaling 1,856,000 parameters. The increased number of convolutional layers is necessary to process the intricate details and distinctions between the 100 classes present in this dataset.
\end{itemize}

The specific choices of model configurations for each dataset were informed by the complexity of the datasets and the total number of classes they included. These configurations are deliberate to ensure that the models are sophisticated enough to capture the relevant features necessary for accurate classification, yet remain efficient in terms of computational resources. The main purpose is to have an optimal balance that maximizes learning capacity for the task at hand while minimizing the risk of forgetting previous tasks.

\subsection{Results and Comparisons}

\subsubsection{Split MNIST}
\begin{figure}[htbp]
\centering
\includegraphics[width=0.9\textwidth]{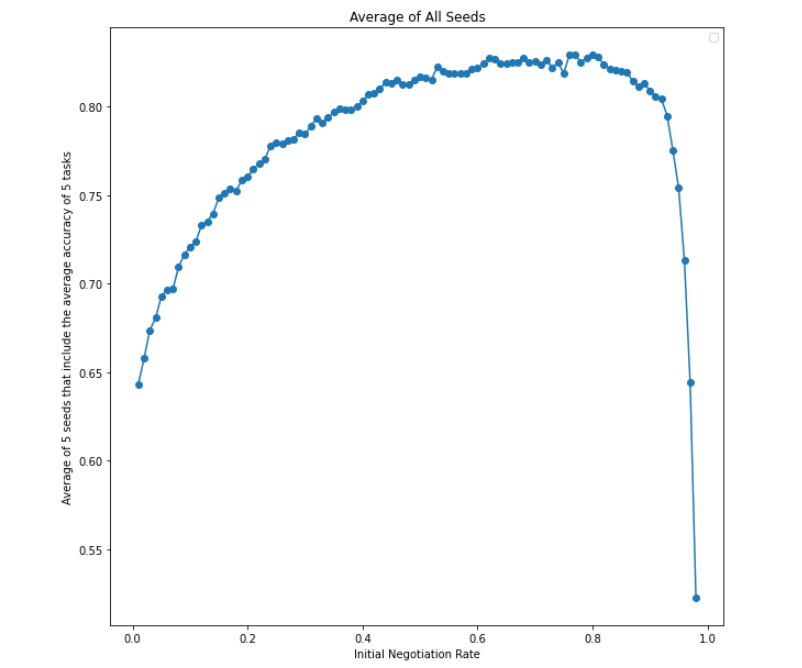}
\caption{Negotiation rate versus the mean of the 5 seeds of the average accuracy of 5 consecutive tasks for Split MNIST. The graph demonstrates the importance of the initial negotiation rate and convolutional layer complexity in achieving high classification accuracy. This graph also tells us that between 0.5 and 0.85, the point where we initiate the negotiations doesn't make much of a difference in average classification accuracy.}
\label{fig:Split_MNIST_graph}
\end{figure}

\textbf{Findings:} The graph in Figure~\ref{fig:Split_MNIST_graph} illustrates an optimal range for the initial negotiation rate that maximizes average accuracy across tasks. As the initial negotiation rate increases, the model's accuracy improves until it reaches a peak, beyond which there is a dramatic decline. This suggests a threshold for the negotiation rate, indicating a point beyond which the model's ability to learn significantly deteriorates. Additionally, due to feature sharing between the Split-MNIST tasks, the initial point of negotiation has a minimal impact on average classification accuracy. 

\textbf{Implications:} These findings underscore the necessity of tuning the initial negotiation rate within an optimal range to balance new knowledge acquisition and previous knowledge retention, mitigating catastrophic forgetting. Furthermore, this balance validates the proposed method's proficiency in managing class incremental learning with minimal loss of prior knowledge. The strategic use of convolutional layers and vector size adjustments further illustrates the model's flexibility in adapting to the learning environment's constraints.

\subsubsection{Split Fashion MNIST}
\begin{figure}[htbp]
\centering
\includegraphics[width=0.9\textwidth]{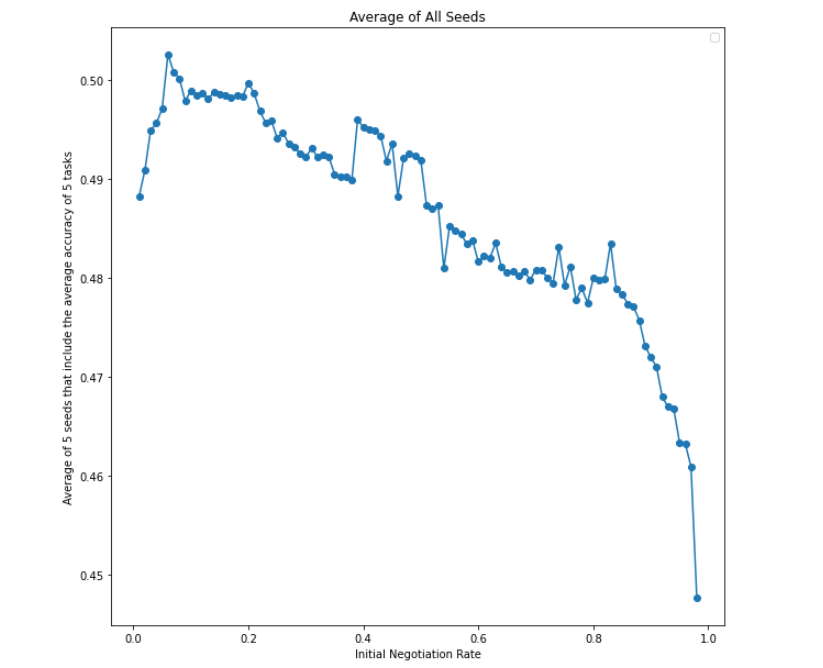}
\caption{Negotiation rate versus the mean of the 5 seeds of the average accuracy of 5 consecutive tasks for Split Fashion MNIST. The graph shows the impact of the initial negotiation rate on the model's accuracy and the compensatory role of vector size in the context of reduced number of convolutional layers.}
\label{fig:Split_Fashion_MNIST_graph}
\end{figure}

\textbf{Findings:} The Figure~\ref{fig:Split_Fashion_MNIST_graph} illustrates the model's performance across various initial negotiation rates. The trend suggests that the accuracy initially fluctuates and then steadily decreases as the negotiation rate increases. This indicates a nuanced relationship between the initial negotiation rate and the model's accuracy. Additionally, the use of a single convolutional layer, despite being less complex, did not hinder the model's ability to learn effectively, which was compensated by increasing the vector size to 64.

\textbf{Implications:} The graph reaffirms the model’s ability to adapt to slightly more complex datasets than Split MNIST like Split Fashion MNIST, highlighting its versatility. It also demonstrates the importance of convolutional layers in the learning process, where even a reduced number can be efficacious given appropriate compensatory measures such as adjusting vector size. The findings suggest the need for careful calibration of the negotiation rate and model architecture to maintain high accuracy levels.

\subsubsection{CIFAR-10}
\begin{figure}[htbp]
\centering
\includegraphics[width=0.9\textwidth]{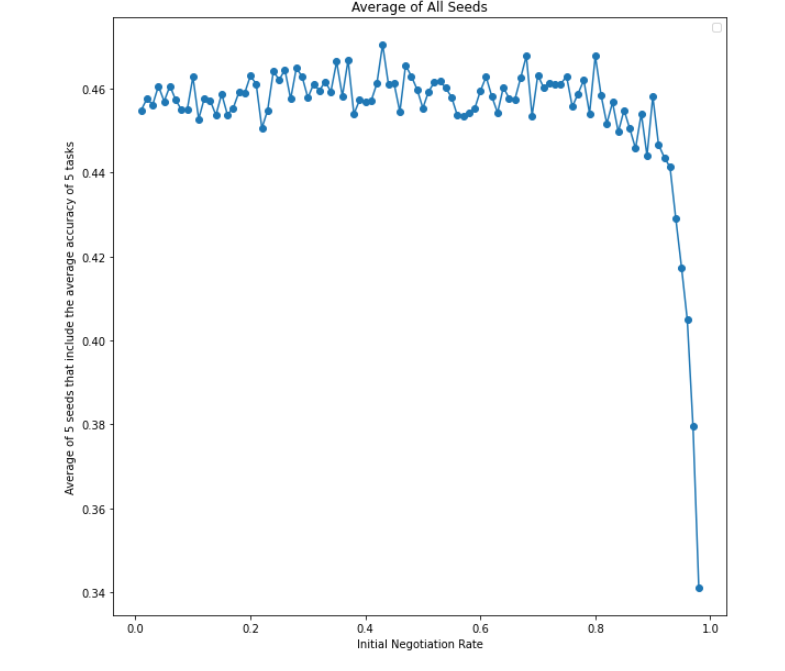}
\caption{Negotiation rate versus the mean of the 5 seeds of the average accuracy of 5 consecutive tasks for CIFAR-10. The graph reflects the impact of the initial negotiation rate on classification accuracy and reveals a critical threshold where the model's performance begins to diminish significantly.}
\label{fig:CIFAR_10_graph}
\end{figure}

\textbf{Findings:} As depicted in Figure~\ref{fig:CIFAR_10_graph}, the model maintains a consistent level of accuracy across a wide range of initial negotiation rates, highlighting its robust learning capability in the context of CIFAR-10's diverse visual concepts. Notably, the accuracy remains relatively stable until the negotiation rate approaches 0.93, beyond which a marked decline is observed.

\textbf{Implications:} This performance trend underscores the model's capacity for handling diverse visual tasks with minimal forgetting, supporting its scalability. The steep drop in accuracy past a negotiation rate of 0.93 suggests a critical limit to the model's flexibility, indicating that too much restriction in the learning process can be detrimental. The negotiation rate's inverse correlation with the model's learning capacity becomes evident, emphasizing the need for a balanced approach to prevent overfitting to previous tasks while still accommodating new knowledge.

\subsubsection{CIFAR-100}
\begin{figure}[htbp]
\centering
\includegraphics[width=0.9\textwidth]{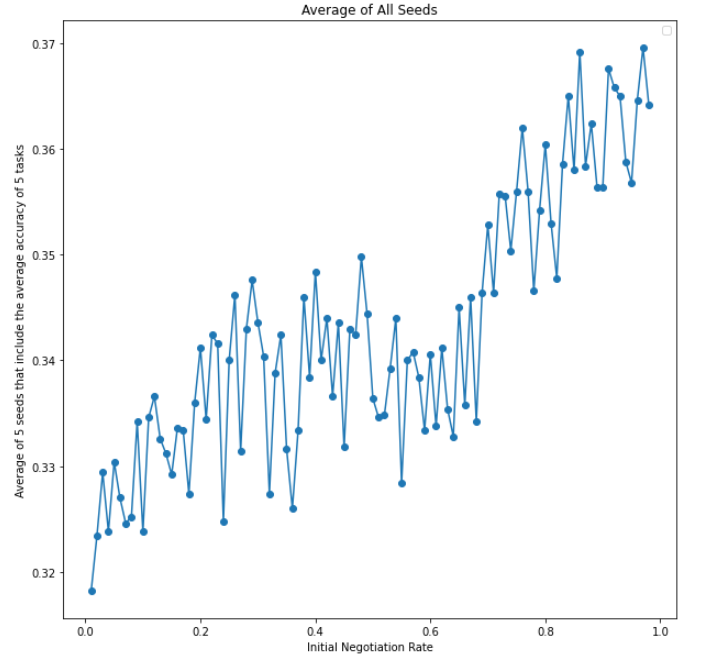}
\caption{Negotiation rate versus the mean of the 5 seeds of the average accuracy of 5 consecutive tasks for CIFAR-100. The graph demonstrates the model's fluctuating performance and underlines the challenges faced due to limited computational resources and model capacity in complex task environments.}
\label{fig:CIFAR_100_graph}
\end{figure}

\textbf{Findings:} The graph in Figure~\ref{fig:CIFAR_100_graph} presents a complex pattern of the model's performance on CIFAR-100, characterized by high variability in accuracy across different initial negotiation rates. This variability suggests that the model experiences significant fluctuations in learning success, which could be attributed to the total number of classes in the CIFAR-100 dataset and the possibility of requiring a larger model.

\textbf{Implications:} The performance curve points to the necessity for models with greater capacity and more computational power to handle the complexity of CIFAR-100 effectively. The graph's pattern emphasizes the need for further investigation into how negotiation rates affect learning in complex scenarios and whether a more sophisticated approach to learning rate adjustment could yield more consistent results. Researchers are encouraged to explore these dynamics with enhanced models to fully capitalize on the potential of advanced machine learning techniques. 

\subsubsection{Comparisons}

The Table~\ref{table:Table_acc} presents a comparative analysis of the average accuracy between a baseline model, which operates without the negotiation mechanism, and a proposed model that incorporates negotiation. In order to avoid randomness of the networks all the experiments are carried out over five different seeds and the average accuracies are shared in the table. The data is segmented by four distinct datasets: MNIST, Fashion MNIST, CIFAR10, and CIFAR100, representing an increasing order of complexity and challenge.

For the MNIST dataset, the baseline model demonstrates an average accuracy of 19.5\%, which is significantly improved upon by the proposed model with negotiation, achieving an impressive 82.3\% accuracy. This substantial increase suggests that the negotiation mechanism plays a critical role in enhancing the model's ability to classify MNIST images correctly.

Similarly, in the Fashion MNIST dataset, the proposed model outperforms the baseline with a notable margin, where the baseline accuracy stands at 19.9\%, and the proposed model achieves 54.8\%. While the increase is not as high as it is with the MNIST dataset, it still indicates a meaningful improvement, especially considering the Fashion MNIST dataset is more complex compared to MNIST dataset.

The trend of improvement continues with the CIFAR10 dataset, where the baseline model's accuracy is 19.3\%, and the proposed model achieves an accuracy of 46.5\%. The CIFAR10 dataset, known for its complexity due to the inclusion of color images, presents a more challenging classification task than the first two. The improvement highlights the proposed model's robustness and its ability to handle more intricate visual information.

The CIFAR100 dataset, which is the most complex among the listed datasets due to its 100 distinct classes, shows the lowest accuracies for both models. However, the proposed model still leads with a significant margin, achieving a 34.9\% accuracy over the baseline model's 13.3\%. Although the accuracies are lower, the relative improvement brought by the negotiation mechanism is considerable. Our approach was limited to negotiation without employing a model of sufficient size for CIFAR 100; nevertheless, the model's performance surpassed that of many studies in the field, aside from those implementing replay mechanisms \cite{sokar2021self,zhou2023deep}.

\begin{table}[H]
\caption{Average accuracy comparison of baseline model and the proposed model on test data.}
\centering
\begin{tabular}{ccccc}
Dataset \ & Baseline Model(without negotiation)\ & Proposed Model(with negotiation) &  \\ \cline{1-4}
MNIST          & 0.195      & 0.823                                  &  \\ 
Fashion MNIST  & 0.199      & 0.548                                  &  \\ 
CIFAR10        & 0.193      & 0.465                                  &  \\ 
CIFAR100       & 0.133      & 0.349                                  &  \\  \cline{1-4}
\end{tabular}
\label{table:Table_acc}
\end{table}

\section{Conclusion}

In this study we have proposed a new method to prevent catastrophic forgetting in class incremental continual learning setups. This method can be easily applied to other scenarios such as domain incremental and task incremental. We initiated our exploration with the class incremental setting due to the fact that it is the most challenging configuration among all available setups. In class incremental learning, the model is tasked with learning new classes sequentially, and this poses several noteworthy challenges. One primary issue is the potential for catastrophic forgetting, where the model may struggle to retain knowledge of previously learned classes when exposed to new ones. This phenomenon can hinder overall performance and necessitates the development of effective strategies to mitigate forgetting and facilitate continual learning across evolving tasks. Additionally, adapting to a growing number of classes demands robust mechanisms for model expansion and efficient utilization of limited resources, adding further complexity to the learning process.

The concept of negotiation in continual learning serves a crucial role in preventing the model from entirely departing its existing knowledge neighborhood when confronted with new tasks. This deliberate constraint is introduced to counteract the potential consequence of unrestricted movement, where the model, if granted absolute freedom to shift, may succumb to the challenge of forgetting previously acquired knowledge. By incorporating negotiation mechanisms, we aim to strike a balance that facilitates adaptability to new tasks while preserving the valuable insights gained from prior learning experiences. This strategic approach seeks to mitigate the risk of catastrophic forgetting and reinforces the model's ability to seamlessly integrate new knowledge into its existing cognitive framework.

 From the results, one can conclude that the integration of a negotiation mechanism within the model's learning paradigm has yielded a substantial improvement in performance across all four datasets. The empirical evidence, as presented in this study, clearly demonstrates the model's enhanced capability to retain knowledge from previously learned tasks while successfully acquiring new information. This balance has been achieved without the need for extensive computational resources, underlining the efficiency of the proposed approach.
Furthermore, the comparative analysis with baseline models reaffirms the superiority of our method. Not only does it reduce the impact of catastrophic forgetting, but it also shows promise for scalability and adaptability in more complex learning environments. This versatility is very important for the advancement of continual learning systems and their application in real-world scenarios. The potential applications of this method extend beyond the academic sphere and into the industry, where systems must continuously learn and evolve without losing the essence of their past experiences. The findings of this study pave the way for future research to explore various dimensions of continual learning, particularly in domains where data streams are non-stationary and tasks are dynamically evolving.

\bibliography{references.bib}

\end{document}